\def\year{2019}\relax
\documentclass[letterpaper]{article} 
\usepackage{aaai19}  
\usepackage{times}  
\usepackage{helvet}  
\usepackage{courier}  
\usepackage[hyphens]{url} 
\usepackage{graphicx}  
\usepackage{appendix}
\usepackage{subfigure}
\usepackage{amsmath}
\usepackage{amsfonts}    
\usepackage{nicefrac} 
\usepackage{natbib}
\usepackage{verbatim}
\usepackage{color}

\DeclareMathOperator*{\argmax}{arg\,max}

\begin{document}
\title{Natural Environment Benchmarks for Reinforcement Learning}
\author{
Amy Zhang \\
McGill University \\ 
Facebook AI Research \\
\texttt{amy.x.zhang@mail.mcgill.ca} \\ 
\And 
Yuxin Wu \\
Facebook AI Research \\
\texttt{yuxinwu@fb.com} \\
\And
Joelle Pineau \\
  McGill University \\ 
  Facebook AI Research \\
  \texttt{jpineau@cs.mcgill.ca} \\ }
  
\maketitle

\begin{abstract}
     While current benchmark reinforcement learning (RL) tasks have been useful to drive progress in the field, they are in many ways poor substitutes for learning with real-world data.  By testing increasingly complex RL algorithms on low-complexity simulation environments, we often end up with brittle RL policies that generalize poorly beyond the very specific domain. To combat this, we propose three new families of benchmark RL domains that contain some of the complexity of the natural world, while still supporting fast and extensive data acquisition. The proposed domains also permit a characterization of generalization through fair train/test separation, and easy comparison and replication of results.  Through this work, we challenge the RL research community to develop more robust algorithms that meet high standards of evaluation. 
     
\end{abstract}

\section{Introduction}
The field of Reinforcement Learning (RL) has exploded in recent years, with over 10K research papers published per year for the last six years~\citep{DBLP:conf/aaai/0002IBPPM18}.  The availability of challenging and widely-used benchmarking domains, such as the Atari Learning Environment~\citep{bellemare13arcade} and MuJoCo~\citep{conf/iros/TodorovET12}, has been a major driver of progress.  By allowing the community to rally around a class of domains, these benchmarks enable fair and easy-to-share comparison of methods, which is useful to properly evaluate progress and ideas.  The widespread use of benchmark datasets has had similar effect, in terms of driving progress, on several other subfields of AI~\citep{lecun-mnisthandwrittendigit-2010,cifar100,Deng09imagenet:a}. In other areas of science, from physics to biology, the use of simulators and models is also common practice.

More recently however, over-reliance on our current RL benchmarks has been called into question~\citep{DBLP:conf/aaai/0002IBPPM18}.  Results showing serious brittleness of methods suggest that either our algorithms are not sufficiently robust, or that our simulators are not sufficiently diverse to induce interesting learned behaviors.   While there is a wealth of work on the former, very few research groups are paying attention to the latter, with the result that we devise increasingly rich algorithms, but continue to test them on synthetic domains of limited complexity which are a poor indicator of real-world performance.

Most benchmarks and datasets used to evaluate machine learning algorithms (excluding RL) consist of data acquired from the real-world, including images, sound, human-written text.  There are cases where synthetic data is considered in early phases of research, but most of the work is done on real-world data.  In contrast, almost all of RL is done with low-complexity synthetic benchmarks.  Of course some work uses robots and other physical systems, but the cost and complexity of data acquisition and platform sharing is prohibitive, and therefore such work can rarely be independently replicated.

The aim of this paper is to explore a new class of RL simulators that incorporate signals acquired from the natural (real) world as part of the state space.  The use of natural signal is motivated by several observations.  First, in comparison to just injecting random noise into the simulator, linking the state to a real-world signal ensures we have more meaningful task characteristics.  Second, by sourcing a component of the state space from the real-world we can achieve fair train/test separation, which is a long-standing challenge for RL\footnote{In simulation domains, RL agents effectively typically train \& test with the same simulator; if the simulator parameters are altered between training and evaluation then it is assumed to be an instance of transfer learning.}.   Yet the tasks we propose remain fast and simple to use; in contrast to other work that might require a common robot infrastructure~\citep{kober13} or animal model~\citep{guez08} or actual plant eco-system~\citep{dietterich18}, our set of tasks requires only a computer and Internet connection.  The domains are easy to install, large quantities of data can be rapidly acquired, and the domains lend themselves to fair evaluations and comparisons.   

In this paper we describe three families of natural environment RL domains, and we provide benchmark performance for several common RL algorithms on these domains.
The three families of domains include two  visual reasoning tasks, where an RL agent is trained to navigate inside natural images to classify images and localize objects, and a variant of the Atari Learning Environment that incorporates natural video in the background.   In the process, we also uncover weaknesses of existing benchmarks that may not be well-recognized in the community.  The primary goal of this work is to encourage the community to tackle RL domains beyond current short-description-length simulators, and to develop methods that are effective and robust in domains with natural conditions.  Some of these new tasks also require RL to achieve higher-order cognition, for example combining the problems of image understanding and task solving.

\section{Motivation}

Consider one of the most widely used simulators for RL benchmarking:  the Atari Learning Environment~\citep{bellemare13arcade}.  In the words of the authors: \textit{ALE is a simple object-oriented framework that allows researchers and hobbyists to develop AI agents for Atari 2600 games. It is built on top of the Atari 2600 emulator Stella.}  The original Atari source code for some of these games is less than 100KB\footnote{\small\url{http://www.atariage.com/2600/programming/index.html}}, the game state evolves in a fully deterministic manner, and there is no further injection of noise to add complexity. Even the core physics engine code for MuJoCo is around 1MB\footnote{Supplied by Emo Todorov.}, which simulates basic physical dynamics of the real world. Thus we argue that the inherent complexity of most ALE games and current physics engines, as defined by the description length of the domain, is trivially small.

Now compare this to a robot that has to operate in the real-world.  The space of perceptual inputs depends on the robot's sensors \& their resolution.  A standard Bumblebee stereo vision camera\footnote{\url{https://www.ptgrey.com/bumblebee2-firewire-stereo-vision-camera-systems}} will generate over 10MB per second.  Now consider that this robot is deployed in a world with zettabytes (=$10^{21}$ bytes) of human-made information\footnote{\url{https://blogs.cisco.com/sp/the-zettabyte-era-officially-begins-how-much-is-that}}, and where each human body may contain upwards of 150 zettabytes\footnote{\url{https://bitesizebio.com/8378/how-much-information-is-stored-in-the-human-genome/}}.  Clearly, RL algorithms have a long way to go before they can tackle the real-world in all its beautiful complexity.

While we strongly support the deployment and evaluation of RL algorithms in real-world domains, there are good reasons to explore protocols that allow replicable evaluation of RL algorithms in a fair and standardized way.  This is the primary goal of this work.  We aim to propose a set of benchmark RL domains that (a) contain some of the complexity of the natural world, (b) support fast and plentiful data acquisition, (c) allow fair train/test separation, and (d) enable easy replication and comparison.

\section{Technical Setting}
In reinforcement learning, an agent interacts with an environment modeled as a Markov Decision Process (MDP)~\citep{bellman1957markovian}, which can be represented by a 6-tuple $(\mathcal{S}, \mathcal{A}, p_0(\mathcal{S}), T, R, \gamma)$, where: 
\begin{itemize}
    \item $\mathcal{S}$ is the set of states,
    \item $\mathcal{A}$ is the set of actions,
    \item $p_0(\mathcal{S})$ is the initial state distribution,
    \item $T(S_{t+1}|S_t,A_t)$ is the probability of transitioning from state $S_t$ to $S_{t+1}$, $S_t,S_{t+1}\in\mathcal{S}$ after action $A_t\in\mathcal{A}$,
    \item $R(r_{t+1}|S_t,A_t)$ is the probability of receiving reward $r_{t+1}\in\mathbb{R}$ after executing action $A_t$ while in state $S_t$,
    \item $\gamma\in[0, 1)$ is the discount factor.
\end{itemize} 

\textbf{Value-based methods} aim to learn the value function of each state or state-action pair of the optimal policy $\pi$. 
We denote the state value function for a particular policy $\pi$ as $V_\pi(s), \forall s\in\mathcal{S}$. The state-action value function is denoted $Q_\pi(s, a), \forall s, a\in (\mathcal{S}, \mathcal{A})$.
In order to find the value functions corresponding to the optimal policy $\pi^*$, we have the update functions:
\begin{equation}
\begin{aligned}
    Q(s_t, a_t) \leftarrow Q(s_t, a_t) + \alpha [r_{t+1}+ \gamma\max_a Q(s_{t+1}, a)\\
    - Q(s_t,a_t)],
\end{aligned}
\end{equation}
\begin{equation}
    V(s_t) = \max_a Q(s_t, a),
\end{equation}
which will converge to optimal $Q^*(S_t,A_t)$ and $V^*(S_t)$.

When learning an estimate $\hat{Q}(\cdot,\cdot|\omega)$ parameterized by $\omega$ of the optimal value function $Q^*$ with temporal-difference methods we use the gradient update:
\begin{align*}
    \omega_{t+1} \leftarrow \omega_t + \alpha [r_{t+1}+\gamma\max_a\hat{Q}(s_{t+1}, a;\omega)\\
     - \hat{Q}(s_t, a_t;\omega)]\nabla_\omega \hat{Q}(s_t, a_t;\omega).
\end{align*}
The optimal policy is found by acting greedily over the optimal value function at each state 
\begin{equation}
\pi^*(s)=\argmax_a Q^*(s,a).
\end{equation}
Learning the state-action value function with this bootstrapping method is called \textit{Q-learning}~\citep{Watkins92q-learning}. Value-based methods are \textit{off-policy} in that they can be trained with samples not taken from the policy being learned. 

\textbf{Policy-based  methods} are methods that directly learn the policy as a parameterized function $\pi_\theta$ rather than learn the value function explicitly, where the parameters of the function are $\mathbf{\theta}$.
Policy gradients use REINFORCE~\citep{Williams92simplestatistical} with the update function
\begin{equation}
\mathbf{\theta}_{t+1}\leftarrow\mathbf{\theta}_t + \alpha G_t \frac{\nabla_\theta\pi(A_t|S_t,\mathbf{\theta}_t)}{\pi(A_t|S_t,\mathbf{\theta}_t)},
\end{equation}
where $\alpha$ is the step size, and $G_t=r_t + \gamma r_{t+1} + \gamma^2 r_{t+2} + ...$ the return. A more general version of REINFORCE uses a baseline $b(S_t)$ to minimize the variance of the update: 
\begin{equation}
\mathbf{\theta}_{t+1}=\mathbf{\theta}_t + \alpha (G_t-b(S_t)) \frac{\nabla_\theta\pi(A_t|S_t,\mathbf{\theta}_t)}{\pi(A_t|S_t,\mathbf{\theta}_t)}.
\end{equation}
This baseline can be an estimate of the state value, learned separately in tabular form or as a parameterized function with weights $\omega$. If the state value function is updated with bootstrapping like in value-based methods, then it is an actor-critic method.

\textbf{Actor-Critic methods} are hybrid value-based and policy-based methods that directly learn both the policy (actor) and the value function (critic)~\citep{NIPS1999_1786}. 
The new update for actor-critic is:
\begin{align}
\mathbf{\theta}_{t+1}&\leftarrow\mathbf{\theta}_t + \alpha (G_t-\hat{V}(S_{t})) \frac{\nabla_\theta\pi(A_t|S_t,\mathbf{\theta}_t)}{\pi(A_t|S_t,\mathbf{\theta}_t)} \\
&=\mathbf{\theta}_t + \alpha (R_{t+1}+\gamma \hat{V}(S_{t+1})-\hat{V}(S_{t})) \frac{\nabla_\theta\pi(A_t|S_t,\mathbf{\theta}_t)}{\pi(A_t|S_t,\mathbf{\theta}_t)},
\end{align}
where $\hat{V}(\cdot)$ is a parameterized estimate of the optimal value function. The corresponding update for $\hat{V}(\cdot)$ is very similar to that in Q-learning~\citep{Watkins92q-learning}:
\begin{equation}
\hat{V}(S_t) \leftarrow \hat{V}(S_t) + \alpha [r_{t+1}+\gamma \hat{V}(S_{t+1})-\hat{V}(S_{t})]
\end{equation}
When learning an estimate $\hat{V}(\cdot|\omega)$ parameterized by $\omega$ of the optimal value function $V^*$ with temporal-difference methods, we use the gradient update:
\begin{equation}
\begin{aligned}
    \omega_{t+1} \leftarrow \omega_t + \alpha [r_{t+1}+\gamma\hat{V}(s_{t+1};\omega) - \hat{V}(s_t;\omega)]\\
    \nabla_\omega\hat{V}(s_t;\omega).
\end{aligned}
\end{equation}

\subsection{Popular RL Algorithms}
\textbf{Advantage Actor Critic (A2C).} ~\cite{2016arXiv160201783M} propose an on-policy method based on actor-critic with several parallel actors which replaces the value estimate with the \textit{advantage} $A_\pi(a, s)=Q_\pi(a,s) - V_\pi(s)$.

\textbf{Actor Critic using Kronecker-Factored Trust Region (ACKTR).} ~\cite{DBLP:journals/corr/abs-1708-05144} uses \textit{trust region optimization} with a \textit{Kronecker-factored approximation (K-FAC)} ~\citep{DBLP:journals/corr/MartensG15} with actor-critic. \textit{Trust region optimization}~\citep{DBLP:journals/corr/SchulmanLMJA15} is an approach where the update is clamped at a maximum learning rate $\eta_{\text{max}}$. \textit{K-FAC} is an invertible approximation of the Fisher information matrix of the neural network representing the policy by block partitioning the matrix according to the layers of the neural network, then approximating these blocks as Kronecker products of smaller matrices. ~\cite{DBLP:journals/corr/MartensG15} show that this approximation is efficient to invert and preserves gradient information. ACKTR is a constrained optimization problem with a constraint that the policy does not move too far in the update, measured with KL-divergence. It also computes steps using the natural gradient direction as opposed to gradient direction. However, computing the exact second derivative is expensive, so \cite{DBLP:journals/corr/abs-1708-05144} instead use K-FAC as an approximation. 

\textbf{Proximal Policy Optimization (PPO).} \cite{2017arXiv170706347S} propose a family of policy gradient methods that also use trust region optimization to clip the size of the gradient and multiple epochs of stochastic gradient ascent for each policy update. PPO uses a penalty to constrain the update to be close to the previous policy.


\textbf{Deep Q-Network (DQN)} \cite{mnih2013playing} modify Q-learning to use a deep neural network with parameters $\omega_t$ to model the state-action value function. The authors introduce a few tricks to stabilize training, mainly using a separate network $Q'$ to compute the target values, which is implemented as an identical neural network but with different parameters $\omega'_t$ copied over from $\omega_t$ at fixed intervals.  The second trick is \textit{experience replay}, or keeping a buffer of prior experience for batch training. The new gradient update to $\omega_t$ is:
\begin{equation}
\begin{aligned}
    \omega_{t+1} \leftarrow \omega_t + \alpha [r_{t+1}+\gamma\max_a Q(s_{t+1},a;\omega') \\
    - Q(s_t,a_t;\omega)]\nabla_\omega Q(s_t,a_t;\omega)
\end{aligned}
\end{equation}

\section{Related Work}
\subsection{Simulation Environments and Benchmarks for RL} There has been many recent proposed simulation engines that try to bridge the gap between simulation and reality by creating more and more realistic but still rendered pixel-level observation spaces~\citep{DBLP:journals/corr/abs-1711-11017,DBLP:journals/corr/abs-1712-05474}.  

The current set of benchmark tasks for RL such as the Atari Learning Environment~\citep{bellemare13arcade} and OpenAI gym~\citep{1802.09464} are primarily composed of deterministic settings. A larger issue is that even in the tasks with larger state spaces, such as the pixel state in Atari, the description of the rules can be modeled in a small instruction set (lines of code or natural language rules). 
The real world is not deterministic, in part because the stochasticity comes from unobserved variables, but is also not directly model-able in a few lines of rules or code. 

\cite{DBLP:journals/corr/DuanCHSA16} released a set of benchmark tasks for continuous RL, pointing out that existing algorithms that work well on discrete tasks~\citep{bellemare13arcade} wouldn't necessarily transfer well to continuous, high dimensional action spaces in the control domain. 

Benchmarks are necessary to evaluate various proposed algorithms and compare them against each other. However, the current suite of available tasks conflate the difficulty of visual comprehension with that of finding an optimal policy, and are a black box for determining how algorithms are actually solving the task. Our results show that visual comprehension is still a difficult task even though we can achieve record scores in Atari on pixel observation. We must take a step back and focus on tasks that can partition the various dimensions along which RL tasks are difficult. 

~\cite{DBLP:conf/aaai/0002IBPPM18} point out issues of reproducibility in deep RL, and we also find that implementations on top of different frameworks~\citep{45381,paszke2017automatic} as built by \cite{baselines} and \cite{pytorchrl} have very different results.

~\cite{NIPS2017_7233} show that the recent improvements in deep RL are not necessarily due to deep neural networks, and that similar improvements can be seen with linear models. They also propose widening the initial state distribution to generate more robust policies. We take this a step further by proposing to widen the state space of the MDP through the introduction of natural signal.  


\subsection{RL for Classical Computer Vision Tasks}
There has been much recent work bridging RL and computer vision (CV), including object detection~\citep{DBLP:journals/corr/CaicedoL15}, object tracking~\citep{8099631,DBLP:journals/corr/ZhangMWW17}, and object classification~\citep{DBLP:journals/corr/abs-1806-07937}. They show it is possible to use RL techniques to perform object localization, and that using RL to localize is a promising direction of research~\citep{DBLP:journals/nature/LeCunBH15}. 
These works show that RL has been successfully applied to visual comprehension tasks, but often using many domain-specific tricks that are not carried over in RL applications. Our work is the first to evaluate on the visual comprehension tasks with state-of-the-art algorithms designed for RL.

There has also been some work applying CV techniques to solve RL problems in robotics~\citep{DBLP:journals/corr/RusuVRHPH16}, games~\citep{mnih2015humanlevel}, and navigation of maps~\cite{DBLP:journals/corr/abs-1804-00168} via pixel-level observation spaces. These typically consist of applying CNNs to RL tasks to process the low-level pixel state, but only with medium-sized convolutional neural networks or fully-connected networks composed of 2-3 layers.

\begin{figure}[t]
    \centering
    \subfigure[Swimmer]{\includegraphics[width=.08\textwidth]{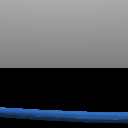}
        \includegraphics[width=.08\textwidth]{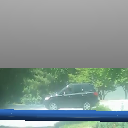}}
    \subfigure[Ant]{\includegraphics[width=.08\textwidth]{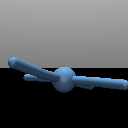}
        \includegraphics[width=.08\textwidth]{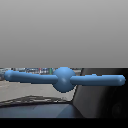}}
    \subfigure[HalfCheetah]{\includegraphics[width=.08\textwidth]{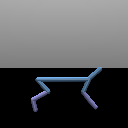}
        \includegraphics[width=.08\textwidth]{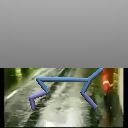}}
    \subfigure[Hopper]{\includegraphics[width=.08\textwidth]{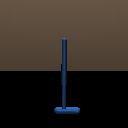}
        \includegraphics[width=.08\textwidth]{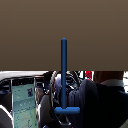}}
            \vskip -0.1in
    \caption{Mujoco frames with original black ground plane (left) and natural video embedded as background in replacement of the ground plane (right).}
    \label{fig:nat_mujoco_frames}
    \vskip -0.15in
\end{figure}

\section{New Benchmark RL Tasks with Natural Signal}
We aim to develop RL benchmarks that capture more of the complexity of the real world, without prohibitive resource and time costs.   We consider three families of tasks, the first two are based on visual reasoning tasks and the third is a variant of existing RL benchmarks.

\subsection{Visual Reasoning using RL}
The first set of proposed tasks consist of gridworld environments overlaid on a natural image. These environments show how we can transform traditionally supervised learning tasks to basic RL navigation tasks with natural signal that requires visual comprehension.  We illustrate this with a few examples (MNIST, CIFAR10 and CIFAR100 for classification; Cityscapes for localization), but the main win here is that we can leverage any existing image dataset. Each of these datasets has a pre-defined train/test split which we respect (train RL agents on training set images; evaluate on test set images) to extract fair generalization measures. These new domains contain several real-world images with natural complexity, and are easily downloadable for easy replication, thus meeting the desiderata outlined in our motivation above. 

\textbf{Agent navigation for image classification.} We propose an image classification task starting with a masked image where the agent starts at a random location on the image. It can unmask windows of the image by moving in one of 4 directions: \{UP, DOWN, LEFT, RIGHT\}. At each timestep it also outputs a probability distribution over possible classes $\mathcal{C}$. The episode ends when the agent correctly classifies the image or a maximum of 20 steps is reached. The agent receives a -0.1 reward at each timestep that it misclassifies the image. The state received at each time step is the full image with unobserved parts masked out.

We evaluate on MNIST~\citep{lecun-mnisthandwrittendigit-2010}, CIFAR10~\citep{cifar}, and CIFAR100~\citep{cifar100}, all of which consist of 60k images, 28x28 and 32x32 respectively, and with 10 classes apiece. MNIST is grayscale (single channel), and CIFAR10 and CIFAR100 are 3 channel RGB.
To scale the difficulty of the problem, we can change the window size $w$ of the agent and maximum number of steps per episode $M$.

\begin{figure}[t]
    \centering
    \subfigure[Breakout]{
        \includegraphics[width=.073\textwidth]{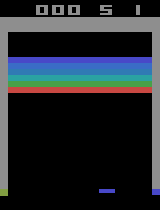}
        \includegraphics[width=.073\textwidth]{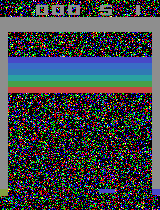}
        \includegraphics[width=.073\textwidth]{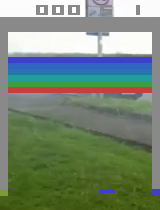}}
    \subfigure[Gravitar]{
        \includegraphics[width=.073\textwidth]{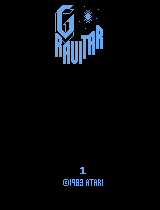}
        \includegraphics[width=.073\textwidth]{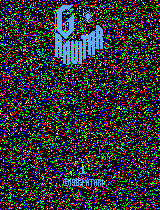}
        \includegraphics[width=.073\textwidth]{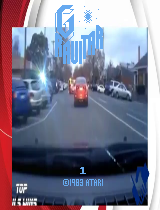}}
        \vskip -0.1in
    \caption{Atari frames, original (left), Gaussian noise (center), and with natural video embedded as background (right).}
    \label{fig:nat_atari_frames}
\end{figure}

\textbf{Agent navigation for object localization.} Given the segmentation mask of an object in an image, the agent has to move to sit on top of the object. There are again 4 possible actions at each timestep, with a time limit of 200 steps. We can further complicate the task with several objects and an additional input of which object class the goal is.

We use the Cityscapes~\citep{Cordts2016Cityscapes} dataset for object detection with a window size $w=10$ which controls the difficulty of the task. The Cityscapes dataset consists of 50k 256x256 images and 30 classes. 
The window size dictates the footprint of the agent. The episode ends if the footprint overlaps with the desired object in the image. The agent is dropped in the center of the image and is given a class label representing a goal object to find and navigate to. The episode ends when the agent is on top of the desired object, for which the environment gives a reward of 1, or if the maximum of 200 steps is reached. There is no penalty for each step in this task -- reward is 0 at each step the agent is not on the desired object.

\begin{figure}[t]
    \centering
    \subfigure{\includegraphics[height=.128\textwidth,trim={.3cm 0 .3cm 0},clip]{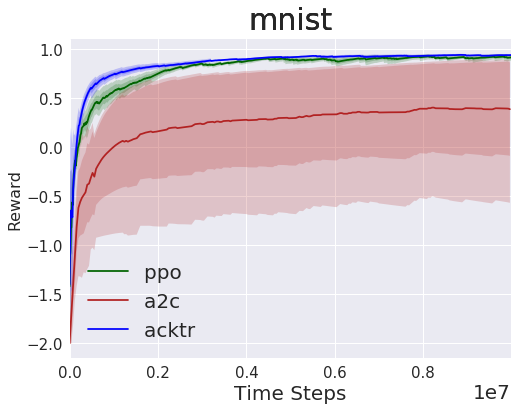}}
    \subfigure{\includegraphics[height=.128\textwidth,trim={1cm 0 .3cm 0},clip]{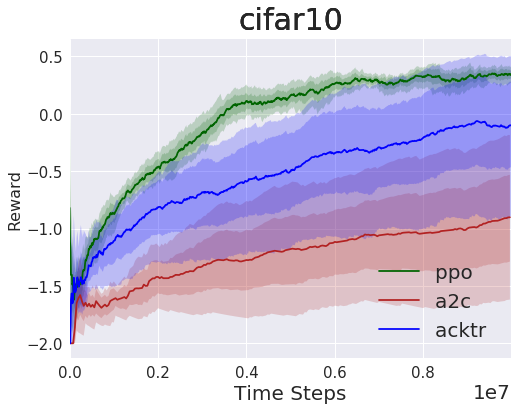}}
    \subfigure{\includegraphics[height=.128\textwidth,trim={1cm 0 .3cm 0},clip]{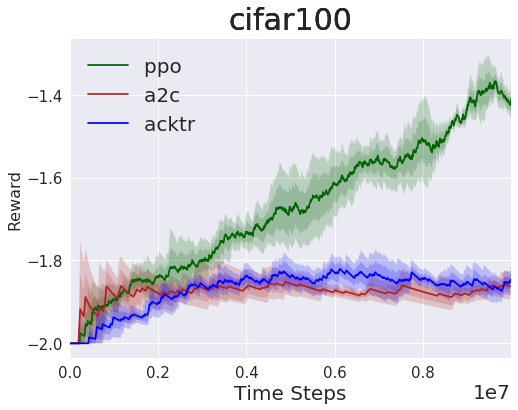}}
        \vskip -0.1in
    \caption{Agent navigation for image classification results. Variance computed across 5 seeds. Note the difference in scale on y-axis.}
    \label{fig:img_clf_results}
    \vskip -0.1in
\end{figure}

\begin{figure}[t]
    \centering
    \subfigure{\includegraphics[height=.128\textwidth,trim={.3cm 0 .3cm 0},clip]{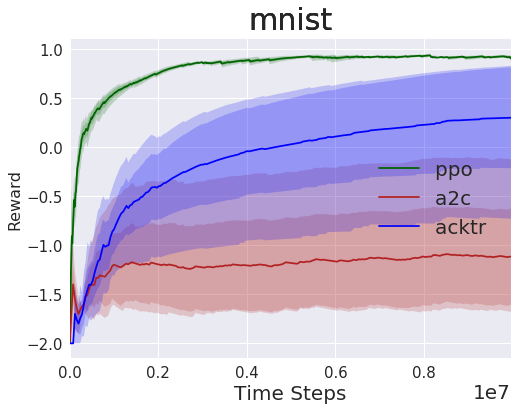}}
    \subfigure{\includegraphics[height=.128\textwidth,trim={1cm 0 .3cm 0},clip]{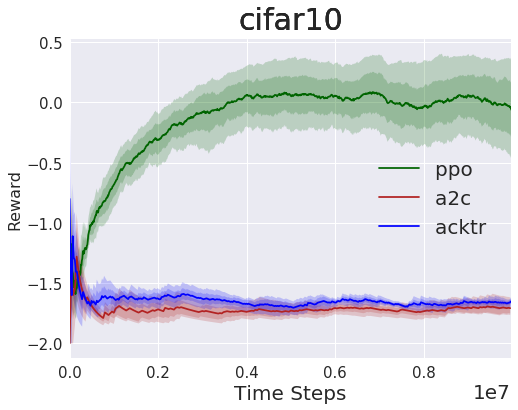}}
    \subfigure{\includegraphics[height=.128\textwidth,trim={1cm 0 .3cm 0},clip]{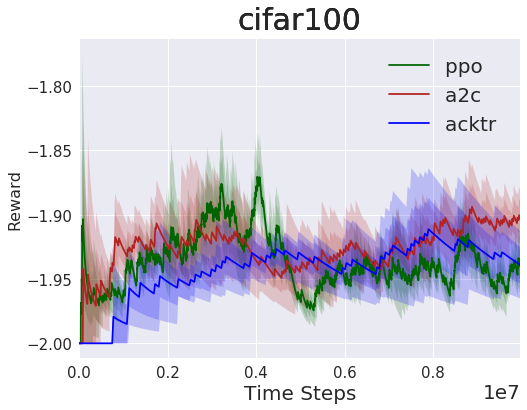}}
    \caption{Agent navigation for image classification results with ResNet-18 trunk. Variance computed across 5 seeds.}
    \label{fig:img_clf_resnet_results}
    \vskip -0.1in
\end{figure}

\subsection{Natural Video RL Benchmarks}

We also propose a modification to existing RL benchmark tasks to incorporate natural signal. Effectively, we take Atari~\citep{bellemare13arcade} tasks from OpenAI gym~\citep{1802.09464} and add natural videos as the background of the observed frames.

We used videos of driving cars from the Kinetics dataset~\citep{DBLP:journals/corr/KayCSZHVVGBNSZ17} and created a mask of the Atari frames by filtering for black pixels $(0, 0, 0 )$, substituting the video frame for the black background. To maintain optical flow we used consecutive frames from randomly chosen videos for the background and randomly sampled from the same set of 840 videos for train and test. 

We do the same for MuJoCo tasks in OpenAI gym~\citep{1802.09464}. The default MuJoCo uses a low-dimensional state space consisting of position and velocity of each joint. Instead, we consider PixelMuJoCo, where the observation space consists of a camera tracking the agent. ~\cite{DBLP:journals/corr/LillicrapHPHETS15} also use a pixel version of MuJoCo and demonstrate similar performance to the low-dimensional version. In our new benchmark, we substitute the floor plane of the PixelMuJoCo tasks (Ant, Swimmer, Hopper, and HalfCheetah) with the same video frames as in the Atari domains. We have included results for PixelMuJoCo but do not include them in our proposed set of benchmarks because we have discovered policies learned for MuJoCo are \textit{open-loop}, and completely ignore the observation input. 

After applying these changes, the state space for these environments drastically increases, and the problem becomes one of visually comprehending the scene in order to attend to the objects corresponding to the game, and ignoring the objects in the video. 
Example frames for Atari and PixelMuJoCo with natural signal can be seen in Figures~\ref{fig:nat_atari_frames}, ~\ref{fig:nat_mujoco_frames}.

\section{Results}
In this section we provide benchmark performance of existing popular RL algorithms on the new proposed domains.
\subsection{Visual Reasoning}
\label{sec:vis_reasoning_results}
For the proposed visual reasoning tasks, we run both a small convolutional neural network (CNN) as commonly used for pixel RL tasks and Resnet-18~\citep{DBLP:journals/corr/HeZRS15} on MNIST~\citep{lecun-mnisthandwrittendigit-2010}, CIFAR10~\citep{cifar}, and Cityscapes~\citep{Cordts2016Cityscapes}. 

The CNN consists of 3 convolutional layers and a fully connected layer, of varying filter sizes and strides to contend with the images from different datasets being different sizes. These layers are interpolated with ReLUs. More detail about model architecture can be found in the Appendix. 

\textbf{Agent navigation for image classification.} Results for the image classification task on MNIST, CIFAR10, and CIFAR100 are found in Figures~\ref{fig:img_clf_results}, \ref{fig:img_clf_resnet_results} for the 3-layer CNN and ResNet-18, respectively. We see that PPO and ACKTR are able to achieve average reward close to 1 on MNIST, which means the agent is able to accurately classify the digit without needing to take many steps.  Performance is worse on CIFAR10 and CIFAR100, as expected, because the datasets consist of more difficult visual concepts. We see the same performance drop across datasets in supervised learning~\citep{DBLP:journals/corr/HeZRS15}. A2C consistently performs worst across all datasets and trunk models. 

More interestingly, we also see performance drop when moving from a 3-layer CNN to ResNet-18 across all 3 datasets. PPO is still able to achieve the same performance on MNIST and CIFAR10, which both have 10 classes, but ACKTR and A2C suffer dramatically. None of the methods work well with ResNet-18 and across 100 classes. This conflates two more difficult problems -- the action space is now $10\times$ larger and there are $10\times$ more concepts to learn. 

\begin{figure}[t]
    \centering
    \subfigure{\includegraphics[width=.23\textwidth]{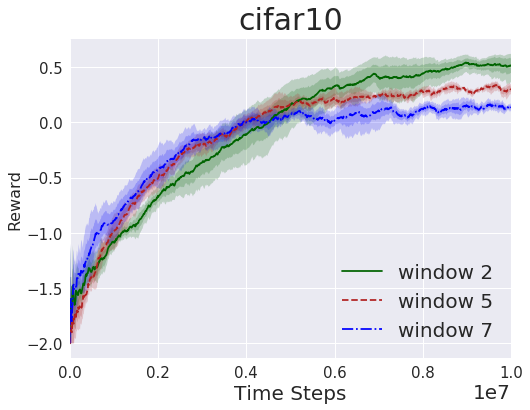}}
    \subfigure{\includegraphics[width=.23\textwidth]{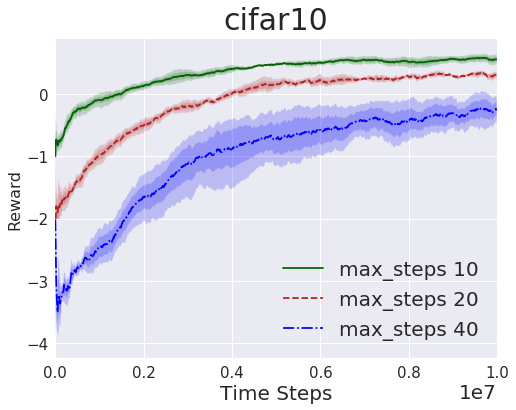}}
        \vskip -0.1in
    \caption{CIFAR10 with PPO, varying window size (left) with fixed maximum number of steps $M=20$ and maximum number of steps per episode (right) with fixed window siz $w=5$.}
    \label{fig:cifar10_vary}
    \vskip -0.1in
\end{figure}

We can alter the difficulty of this task along two dimensions -- varying the window size $w$ of the agent, or the maximum number of steps per episode $M$. In experiments, we try values of $w\in[2, 5, 7]$ and $M\in[10, 20, 40]$. Results are in Figure~\ref{fig:cifar10_vary}. Performance increases with fewer number of steps, which corresponds to more immediate rewards and therefore an easier RL task. Initially, a larger window size performs better, as expected, but as training continues the smaller window $w=2$ dominates. 

We see that the large models that have been successful in SL for object classification suffer in RL tasks. However, accurate object classification and localization are necessary components of how humans solve many tasks, and how we expect RL algorithms to learn to solve tasks. Here we show by isolating the visual comprehension component of common RL tasks how well current state-of-the-art RL algorithms actually perform, and that simple plug and play of successful SL vision models does not give us the same gains when applied in an RL framework.  We thus expect this new family of tasks to spur new innovation in RL algorithms.

\begin{figure}[t]
    \centering
    \subfigure{\includegraphics[width=.23\textwidth]{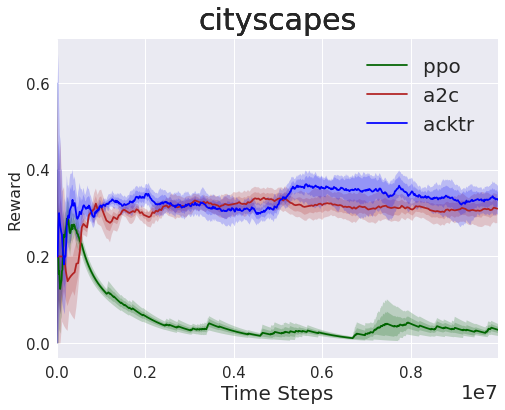}}
    \subfigure{\includegraphics[width=.23\textwidth]{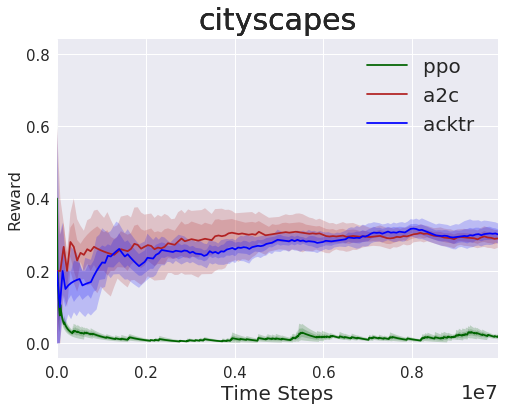}}
        \vskip -0.1in
    \caption{Agent navigation for image detection results. 3-layer CNN (left), ResNet-18 (right).}
    \label{fig:img_det_results}
    \vskip -0.15in
\end{figure}

\textbf{Agent navigation for object localization.} Results for the object detection task on Cityscapes is found in Figure~\ref{fig:img_det_results}. Object detection is a much more difficult task with again a small drop in performance when moving from the 3-layer CNN trunk to ResNet-18. 

Here we see that PPO completely fails to learn, whereas it beat out both  A2C and ACKTR in the classification task. But both A2C and ACKTR are not able to navigate to the desired object in the image more than 40\% of the time. 

Both of these vision tasks demonstrate what others have also found~\citep{rlblogpost,NIPS2017_7233} -- that deep models do not perform well in the RL framework in the same way they do in the SL framework.  Clearly, this opens up many interesting directions for future research.

\subsection{Natural Signal in RL Tasks}
For both Atari~\citep{bellemare13arcade} and PixelMuJoCo~\citep{1802.09464} we follow the preprocessing done in~\cite{mnih2015humanlevel} and resize the frames to $84\times 84$, convert to grayscale, and perform frame skipping and sticky actions for 4 consecutive steps. We also stack four consecutive frames for each observation.
For algorithm implementations we use OpenAI Baselines~\citep{baselines} and Ilya Kostrikov's implementation~\citep{pytorchrl}.

As baseline, we compare with the original Atari and PixelMuJoCo tasks with static black background\footnote{Refer to Fig. \ref{fig:nat_atari_frames} for visualizations of what the original and modified Atari games look like.}. 
\begin{figure}[h]
    \vskip -0.1in
    \centering
    \subfigure{\includegraphics[width=.22\textwidth]{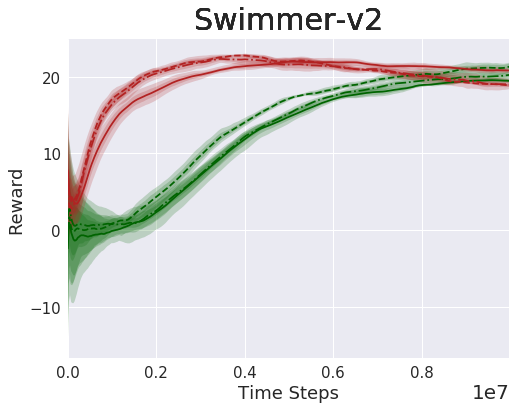}}
    \subfigure{\includegraphics[width=.22\textwidth]{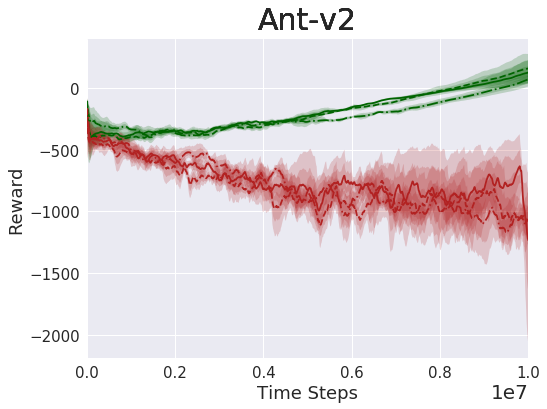}}
    \subfigure{\includegraphics[width=.22\textwidth]{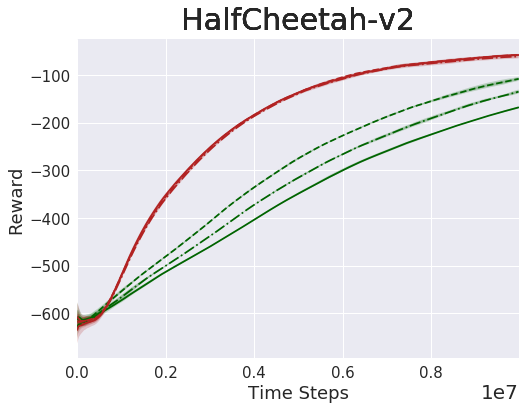}}
    \subfigure{\includegraphics[width=.22\textwidth]{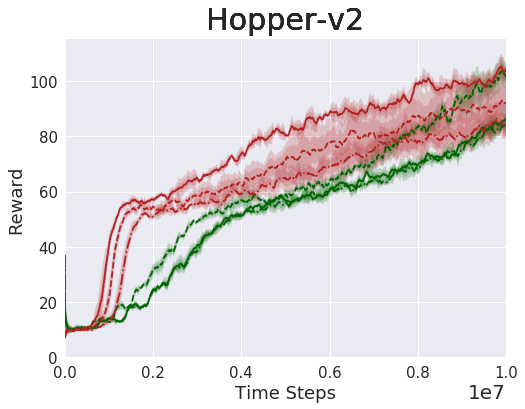}}
         \vskip -0.1in
    \subfigure{\includegraphics[width=1\columnwidth]{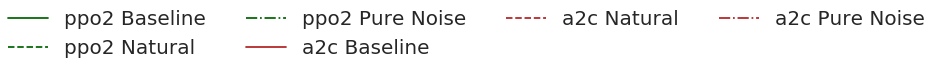}}
        \vskip -0.1in
    \caption{Natural signal in PixelMuJoCo results, using \cite{baselines} code implementation. Variance computed across 5 seeds.}
    \label{fig:nat_mujoco_results}
    \vskip -0.1in
\end{figure}

\textbf{PixelMuJoCo.} For PixelMuJoCo, we evaluate on Hopper, Swimmer, Ant, and HalfCheetah. There are results reported by \cite{DBLP:journals/corr/LillicrapHPHETS15} for PixelMuJoCo with DDPG, but the rewards are not directly comparable, and the pixel environment they use is different from the renderer provided in OpenAI gym based on the visualizations in the paper, and has not been opensourced. 
PixelMuJoCo results are in Figure~\ref{fig:nat_mujoco_results}. We see similar performance across baseline and natural, with small performance gaps apparently especially in HalfCheetah and Hopper. We suspect this is actually caused by the policy falling into a local optima where it ignores the observed state entirely -- the fact that ~\cite{DBLP:journals/corr/LillicrapHPHETS15} also report similar results for the low-dimensional state space and pixel space for MuJoCo tasks also points to this conclusion.

To test our hypothesis, we replace the observation with Gaussian noise. Results are shown in Figure~\ref{fig:nat_mujoco_results} as Pure Noise. Even in the case where there is \textit{no information} returned in the observation---it is pure iid Gaussian noise---PPO and A2C are able to learn as good a policy as when actual state information is provided in the observation.
Our results show that current RL algorithms are solving MuJoCo tasks as an \textit{open-loop control system}, where it completely ignores the output when deciding the next action.  These results suggest that MuJoCo is perhaps not a strong benchmark for RL algorithms, and a good test for open-control policies is substituting the observation with pure noise.  

 \begin{figure*}[t]
     \centering
     \subfigure{\includegraphics[width=.24\textwidth]{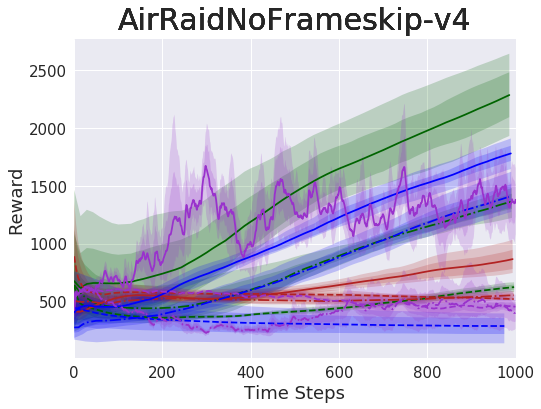}}
     \subfigure{\includegraphics[width=.24\textwidth]{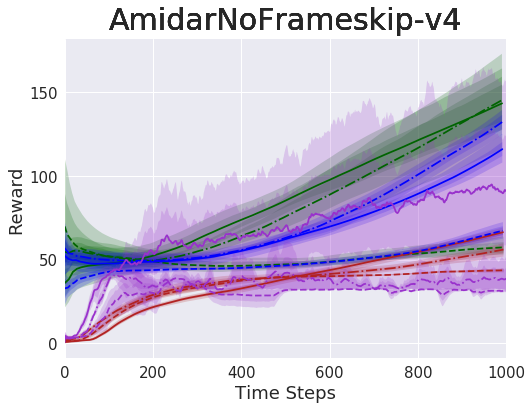}}
     \subfigure{\includegraphics[width=.24\textwidth]{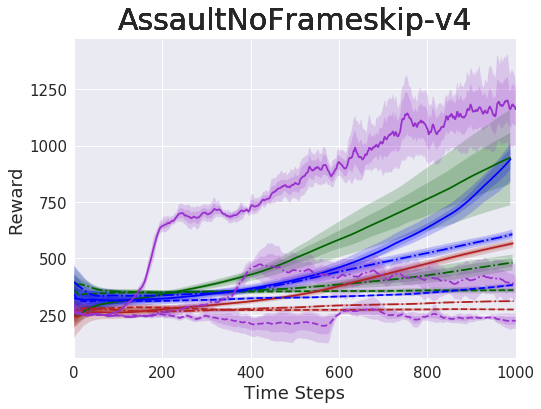}}
     \subfigure{\includegraphics[width=.24\textwidth]{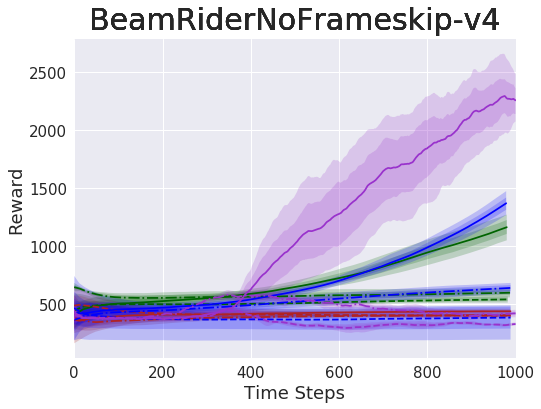}}
     \vskip -0.1in
     \subfigure{\includegraphics[width=1\textwidth]{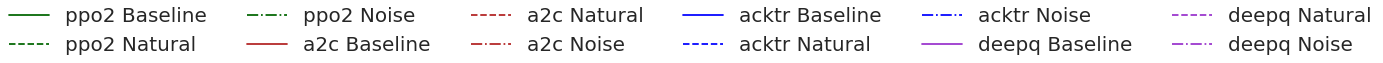}}
         \vskip -0.1in
\caption{Atari frames with baseline, Gaussian noise, and natural video embedded as background, using \cite{baselines} code implementation. Variance computed across 5 seeds.}
     \label{fig:natnoise_atari_results}
\vskip -0.1in
 \end{figure*}
 
\textbf{Atari.} For Atari, we selected 16 environments (mainly ones with black backgrounds for ease of filtering) and evaluated PPO, ACKTR, A2C, and DQN on both the default environment and with injected video frames. Full results can be seen in Fig.~\ref{fig:nat_atari_results} in the Appendix, we have only included 4 environments in the main paper ( Fig.~\ref{fig:natnoise_atari_results}) because of space constraints. We see much larger gaps in performance for many games, showing that visual comprehension is more crucial in these environments. The addition of natural noise with visual flow causes the policy to completely fail in some games, while causing only a small drop in performance in others. In these cases it is again possible that the policy is treating the task as an open-loop control problem.

To see how much the performance difference is caused by the addition of \textit{natural signal} as opposed to just changing these environments to no longer be deterministic, we also evaluate on a few Atari environments where the background is replaced with random i.i.d. Gaussian noise changing from frame to frame (Figure~\ref{fig:natnoise_atari_results}).  We see with all four games that we have best performance with the original static black background (Baseline), similar to reduced performance with Gaussian noise (Noise), and worst performance with video (Natural). However, the performance difference varies. In Amidar, performance with random noise is very close to baseline for ACKTR and PPO, but suffers a massive drop with natural signal. In Beamrider, the algorithms all fail to obtain good policies with any addition of random noise or natural signal. 

We see that the Atari tasks are complex enough, or require different enough behavior in varying states that the policy cannot just ignore the observation state, and instead learns to parse the observation to try to obtain a good policy. In most games it is able to do this with static background, suffers with random noise background, and fails completely with natural signal background. 

\section{Discussion}
We have proposed three new families of benchmark tasks to dissect performance of RL algorithms. The first two are domains that test visual comprehension by bringing traditional supervised learning tasks into the RL framework. In the process, we have  shown that naive plug and play of successful vision models fails in the RL setting. This suggests that the end-to-end frameworks espoused for RL currently are not successful at implicitly learning visual comprehension.

The third family of tasks call for evaluating RL algorithms via incorporating signal from the natural world, by injecting frames from natural video into current RL benchmarks. We have shown that performance deteriorates drastically in this setting across several state-of-the-art RL optimization algorithms and trunk models. With this new set of tasks, we call for new algorithm development to be more robust to natural noise. We also observe that state-of-the-art performance on the PixelMuJoCo domain can be achieved with an open-loop policy, making it an odd choice for an RL benchmark. Based on these results, we have also proposed replacing the observation with pure noise as a test for open-loop policies.

As a side note, we note that we were able to achieve the same results reported by ~\cite{baselines} for some of the games but not all with the default hyperparameters provided and their code, and also saw large differences in performance when comparing results using  \cite{pytorchrl} vs. \cite{baselines} implementations. Results across 16 Atari games are shown in Figures~\ref{fig:nat_atari_results} and \ref{fig:pytorch_nat_atari_results} in the Appendix.

The first set of tasks in object recognition and localization are tests of how well models can learn visual comprehension in an RL setting. Only once we have achieved good results in those tasks can we move onto tasks with more difficult dynamics and be sure that the performance is caused by visual comprehension as opposed to memorizing trajectories. 

Beyond piping natural signal into the state space through the observations, another type of noise in the real world is noise in the action effects and dynamics. Transitions from one state to the next exhibit noise from imperfect actuators and sensors. It is still an open question how we can inject natural dynamics signal into simulated environments.

 \bibliography{nat_rl}
 \bibliographystyle{aaai}

 \appendix
 \newpage

 \section{Implementation}

\subsection{Model Architectures} \label{app:architectures}
\small\begin{verbatim}
if dataset == 'mnist':
    self.main = nn.Sequential(
        nn.Conv2d(num_inputs, 10, 5, stride=2),
        nn.ReLU(),
        nn.Conv2d(10, 20, 5),
        nn.ReLU(),
        nn.Conv2d(20, 10, 5),
        nn.ReLU(),
        Flatten(),
        nn.Linear(360, 512),
        nn.ReLU()
    )
elif dataset == 'cifar10':
    self.main = nn.Sequential(
        nn.Conv2d(num_inputs, 6, 5, stride=2),
        nn.ReLU(),
        nn.Conv2d(6, 16, 5),
        nn.ReLU(),
        nn.Conv2d(16, 6, 5),
        nn.ReLU(),
        Flatten(),
        nn.Linear(216, 512),
        nn.ReLU()
    )
elif dataset == 'cityscapes':
    self.main = nn.Sequential(
        nn.Conv2d(num_inputs, 32, 8, stride=7),
        nn.ReLU(),
        nn.Conv2d(32, 64, 4, stride=4),
        nn.ReLU(),
        nn.Conv2d(64, 32, 3, stride=1),
        nn.ReLU(),
        Flatten(),
        nn.Linear(32 * 7 * 7, 512),
        nn.ReLU()
    )
else:  # MuJoCo and Atari
    self.main = nn.Sequential(
        nn.Conv2d(num_inputs, 32, 8, stride=4),
        nn.ReLU(),
        nn.Conv2d(32, 64, 4, stride=2),
        nn.ReLU(),
        nn.Conv2d(64, 32, 3, stride=1),
        nn.ReLU(),
        Flatten(),
        nn.Linear(32 * 7 * 7, 512),
        nn.ReLU()
    )
\end{verbatim}
\normalsize

\subsection{Hyperparameters}
Hyperparameters were kept constant for all MuJoCo and Atari tasks.

\section{Sample Environment Frames}
 \label{app:frames}
 \begin{figure*}[b]
     \centering
     \subfigure[AirRaid]{\includegraphics[width=.12\textwidth]{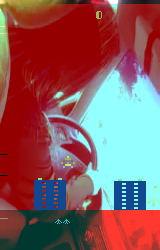}}
     \subfigure[Alien]{\includegraphics[width=.12\textwidth]{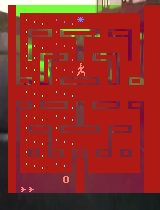}}
     \subfigure[Amidar]{\includegraphics[width=.12\textwidth]{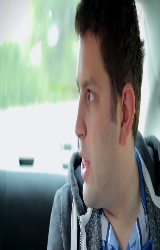}}
     \subfigure[Assault]{\includegraphics[width=.12\textwidth]{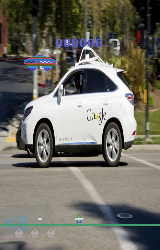}}
     \subfigure[Asteroids]{\includegraphics[width=.12\textwidth]{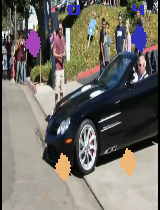}}
     \subfigure[BeamRider]{\includegraphics[width=.12\textwidth]{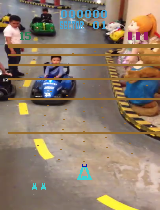}} \\
     \subfigure[Carnival]{\includegraphics[width=.12\textwidth]{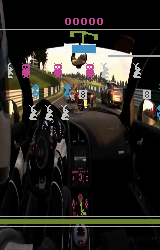}}
     \subfigure[Centipede]{\includegraphics[width=.12\textwidth]{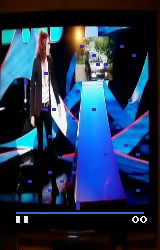}}
     \subfigure[DemonAttack]{\includegraphics[width=.12\textwidth]{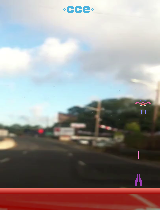}}
     \subfigure[Phoenix]{\includegraphics[width=.12\textwidth]{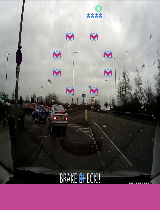}}
     \subfigure[SpaceInvaders]{\includegraphics[width=.12\textwidth]{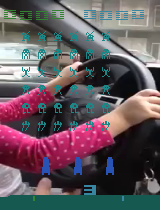}}
     \subfigure[StarGunner]{\includegraphics[width=.12\textwidth]{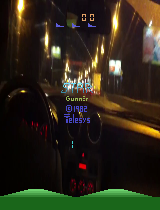}}
     \caption{More Atari frames with natural video embedded as background.}
     \label{fig:app_nat_atari_frames}
 \end{figure*}

 \section{More Results}
 \label{app:results}

\begin{figure*}[h]
    \centering
    \subfigure{\includegraphics[width=.22\textwidth]{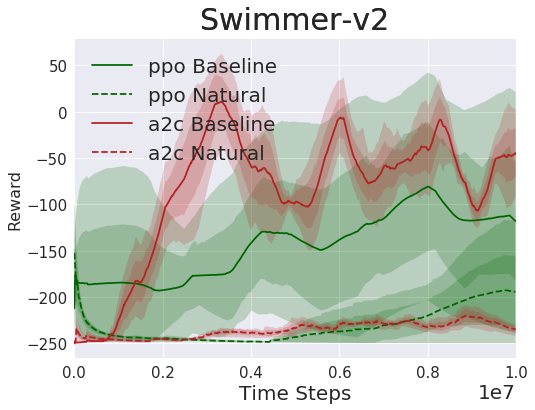}}
    \subfigure{\includegraphics[width=.22\textwidth]{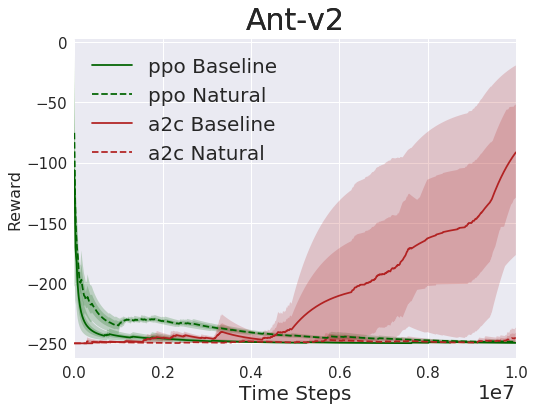}}
    \subfigure{\includegraphics[width=.22\textwidth]{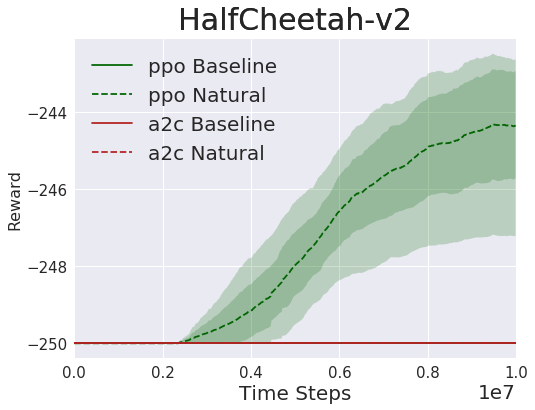}}
    \subfigure{\includegraphics[width=.22\textwidth]{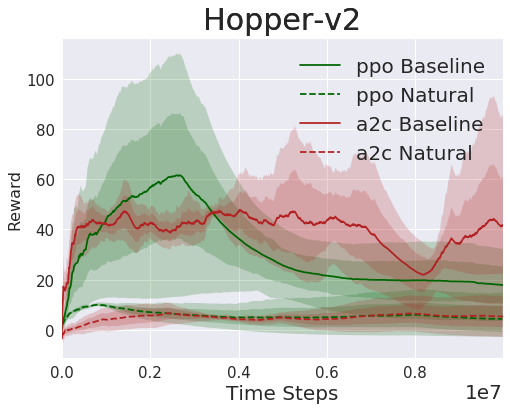}}
    \caption{Natural signal in PixelMuJoCo results, using \cite{pytorchrl} code implementation. Variance computed across 5 seeds.}
    \label{fig:pytorchnat_mujoco_results}
\end{figure*}

 \begin{figure*}[t]
     \centering
     \subfigure{\includegraphics[width=.24\textwidth]{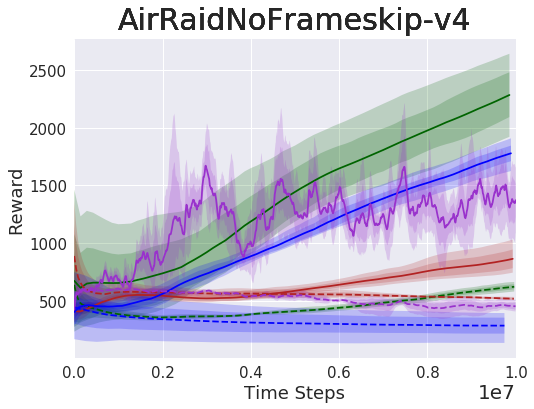}}
     \subfigure{\includegraphics[width=.24\textwidth]{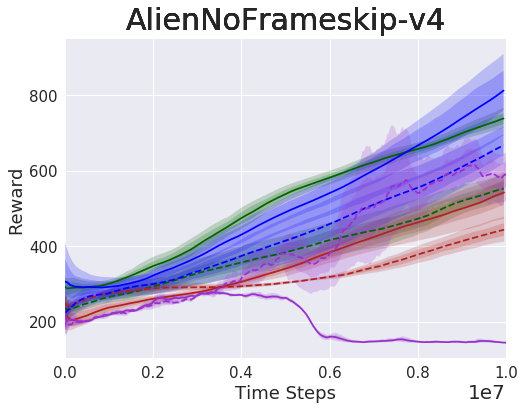}}
     \subfigure{\includegraphics[width=.24\textwidth]{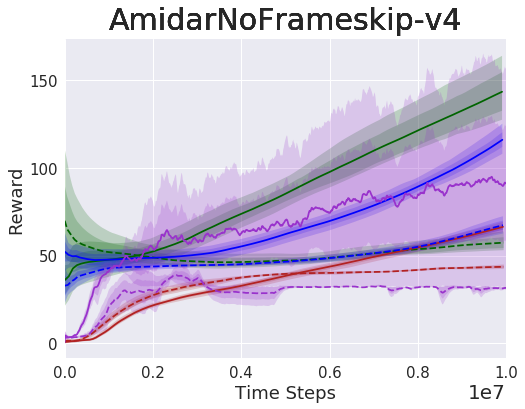}}
     \subfigure{\includegraphics[width=.24\textwidth]{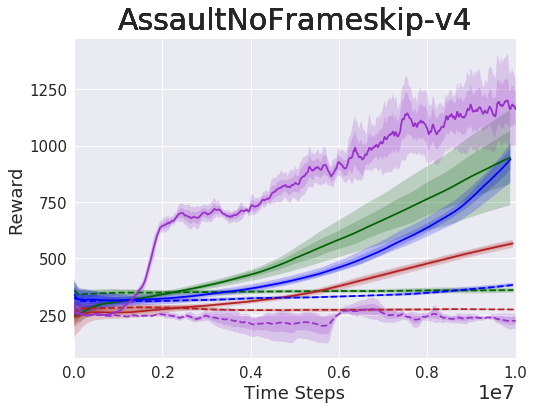}}
     \subfigure{\includegraphics[width=.24\textwidth]{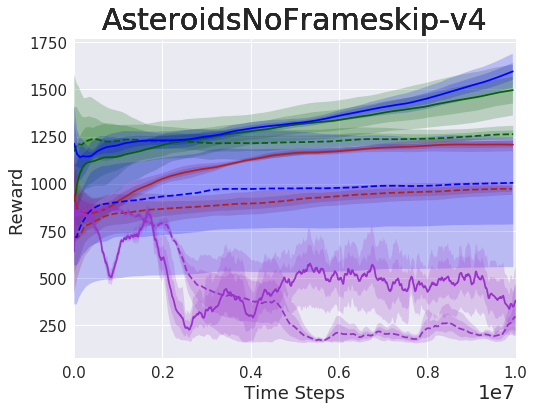}}
     \subfigure{\includegraphics[width=.252\textwidth]{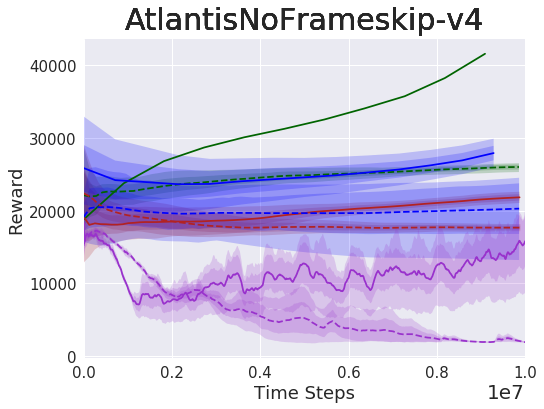}}
     \subfigure{\includegraphics[width=.24\textwidth]{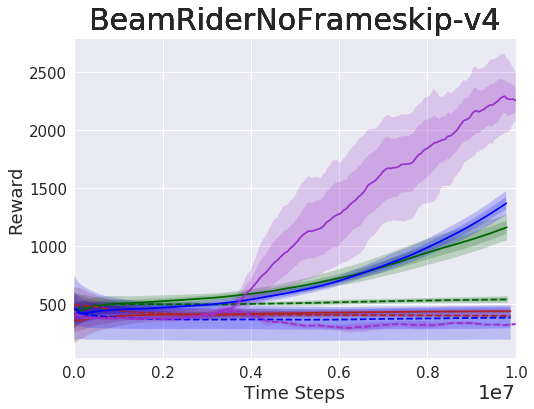}}
     \subfigure{\includegraphics[width=.239\textwidth]{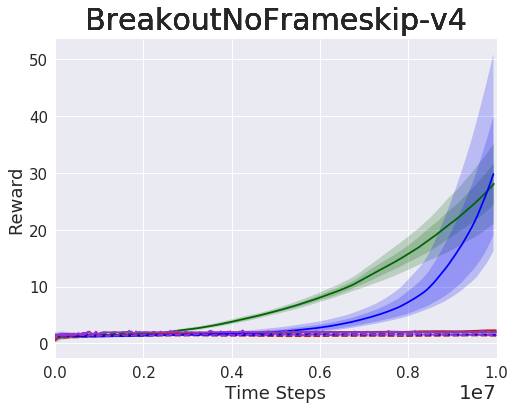}}
     \subfigure{\includegraphics[width=.24\textwidth]{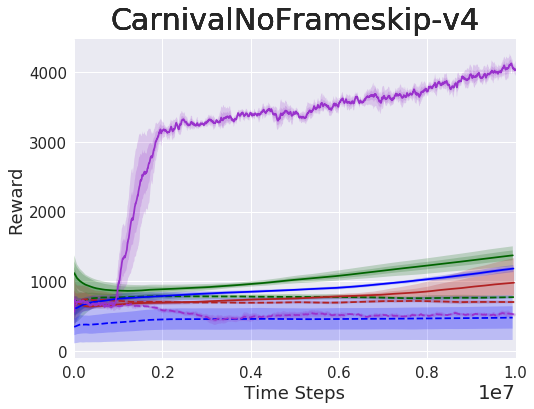}}
     \subfigure{\includegraphics[width=.24\textwidth]{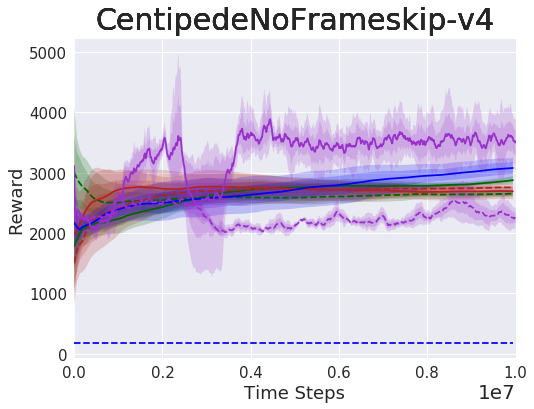}}
     \subfigure{\includegraphics[width=.24\textwidth]{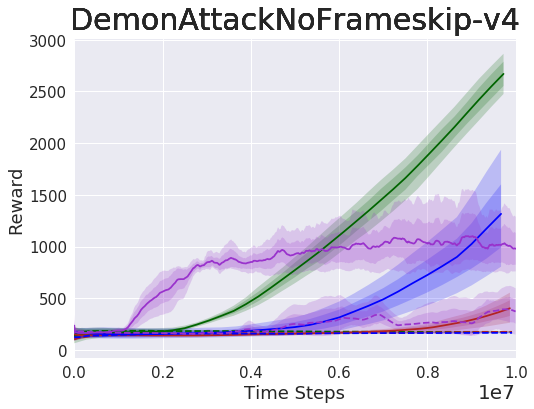}}
     \subfigure{\includegraphics[width=.24\textwidth]{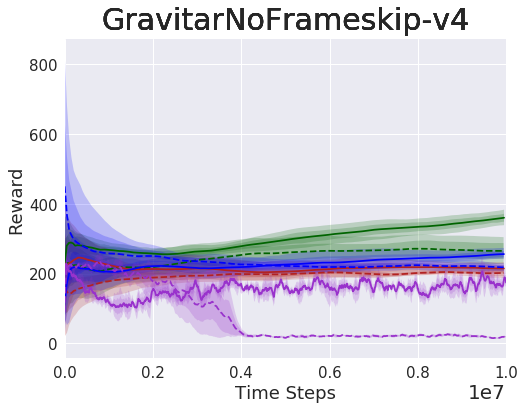}}
     \subfigure{\includegraphics[width=.252\textwidth]{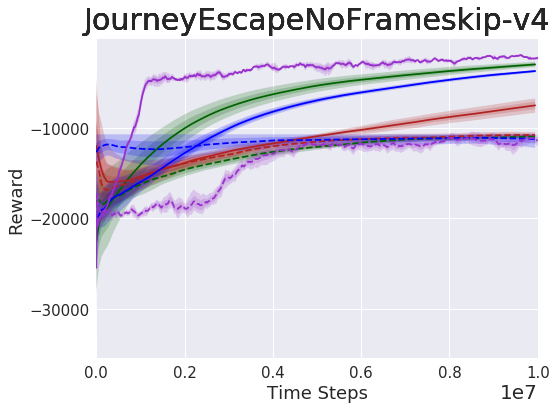}}
     \subfigure{\includegraphics[width=.24\textwidth]{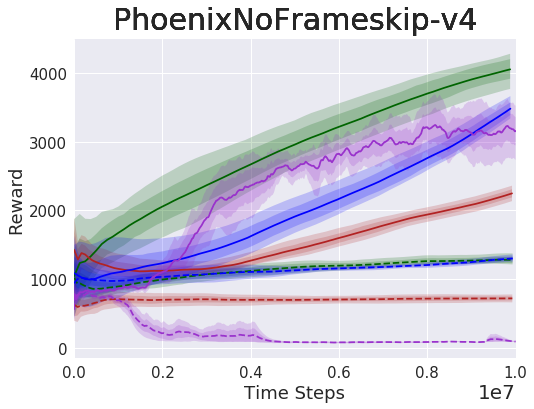}} 
     \subfigure{\includegraphics[width=.24\textwidth]{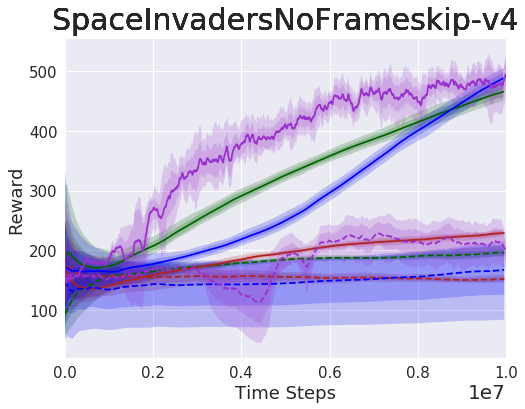}}
     \subfigure{\includegraphics[width=.24\textwidth]{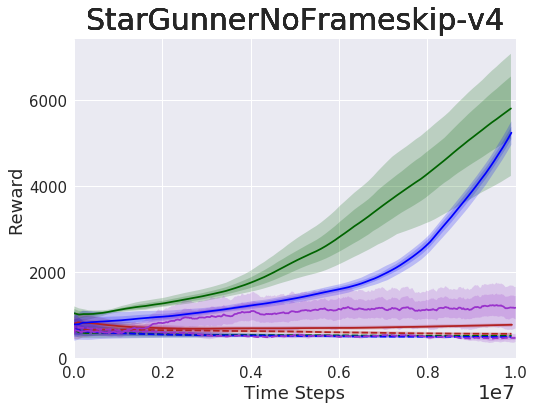}}
          \vskip -0.1in
     \subfigure{\includegraphics[width=1\textwidth]{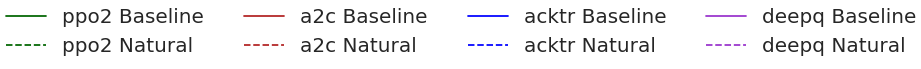}}
     \caption{Atari frames with natural video embedded as background, using \cite{baselines} code implementation. Variance computed across 5 seeds.}
     \label{fig:nat_atari_results}
 \end{figure*}
 
\begin{figure*}[t]
     \centering
     \subfigure{\includegraphics[width=.24\textwidth]{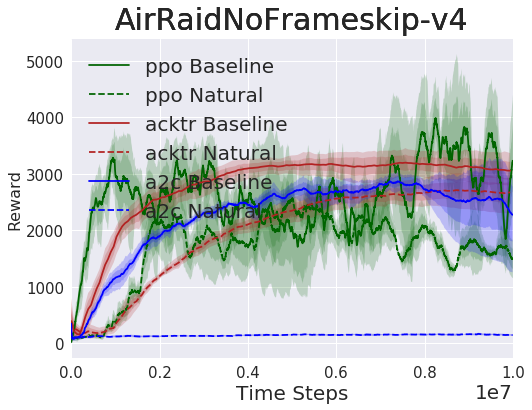}}
     \subfigure{\includegraphics[width=.24\textwidth]{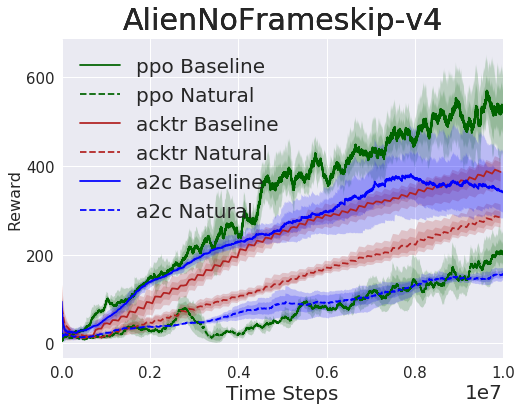}}
     \subfigure{\includegraphics[width=.24\textwidth]{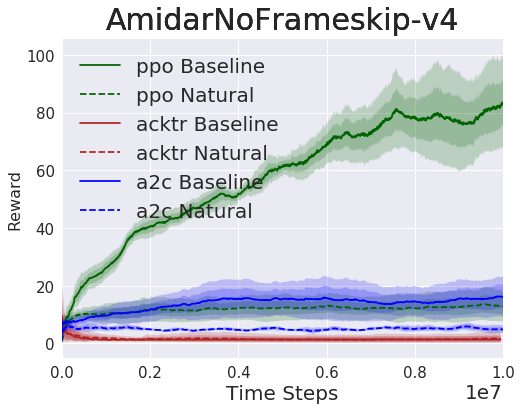}}
     \subfigure{\includegraphics[width=.24\textwidth]{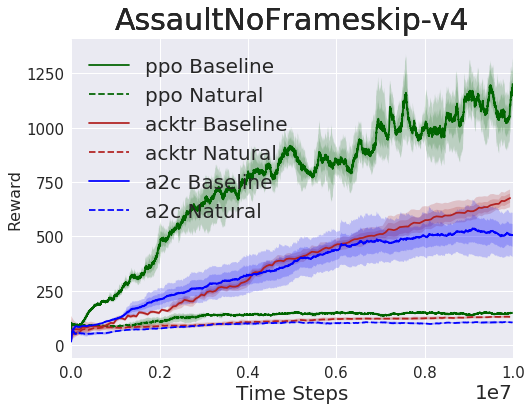}}
     \subfigure{\includegraphics[width=.24\textwidth]{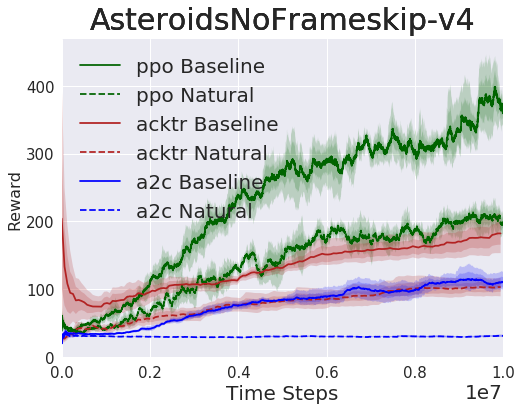}}
     \subfigure{\includegraphics[width=.252\textwidth]{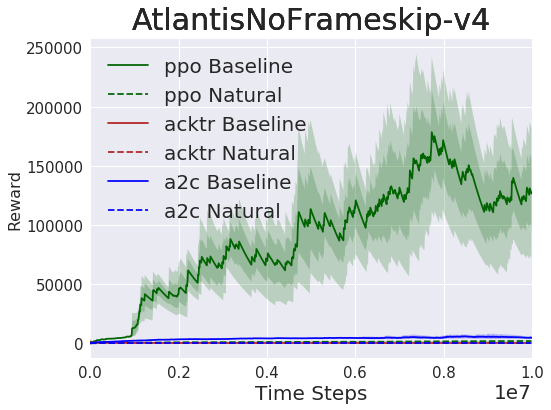}}
     \subfigure{\includegraphics[width=.24\textwidth]{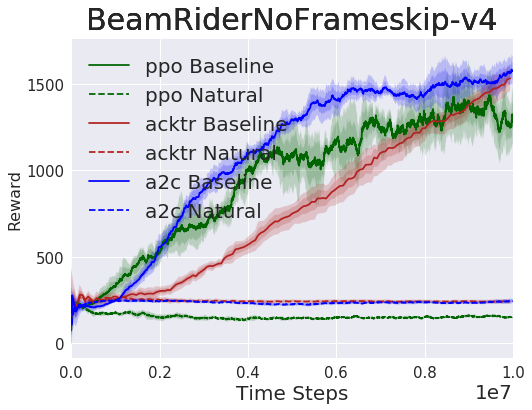}}
     \subfigure{\includegraphics[width=.239\textwidth]{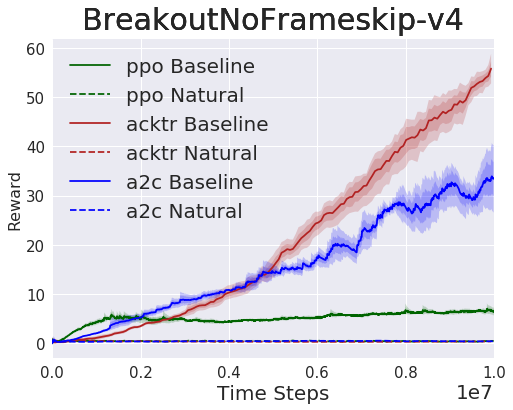}}
     \subfigure{\includegraphics[width=.24\textwidth]{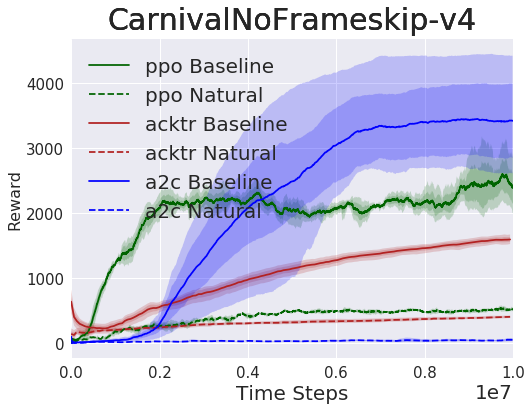}}
     \subfigure{\includegraphics[width=.24\textwidth]{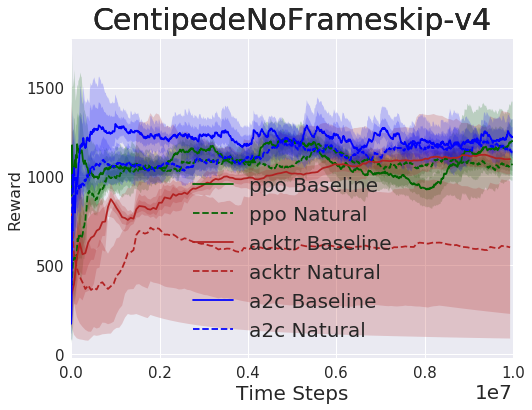}}
     \subfigure{\includegraphics[width=.24\textwidth]{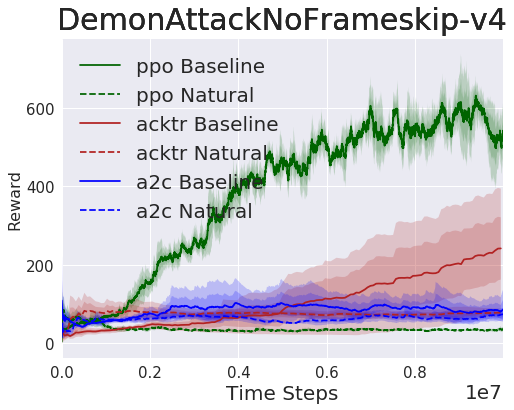}}
     \subfigure{\includegraphics[width=.24\textwidth]{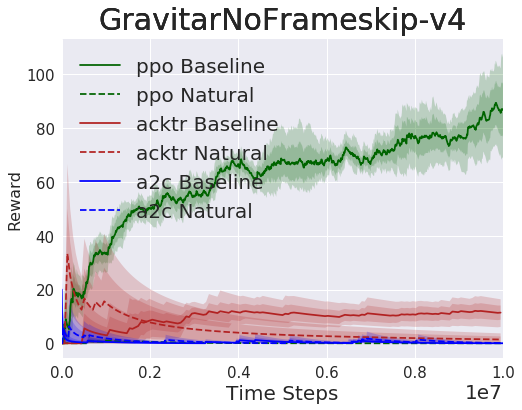}}
     \subfigure{\includegraphics[width=.252\textwidth]{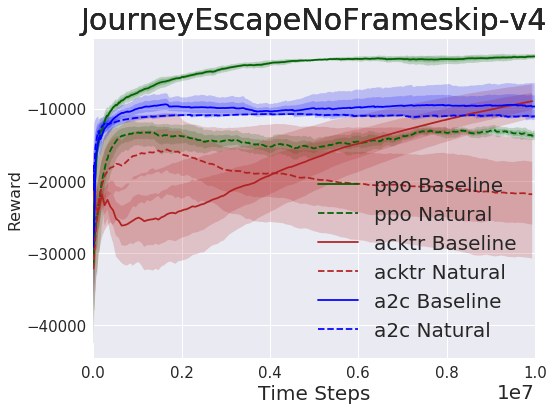}}
     \subfigure{\includegraphics[width=.24\textwidth]{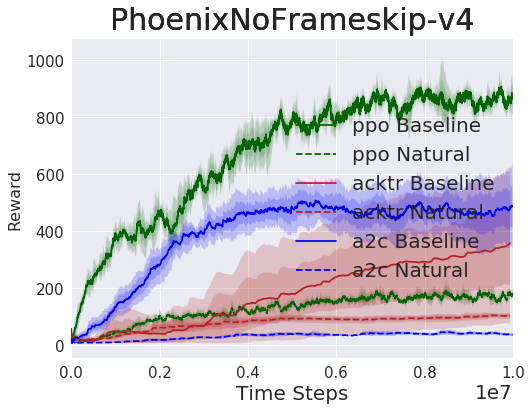}} 
     \subfigure{\includegraphics[width=.24\textwidth]{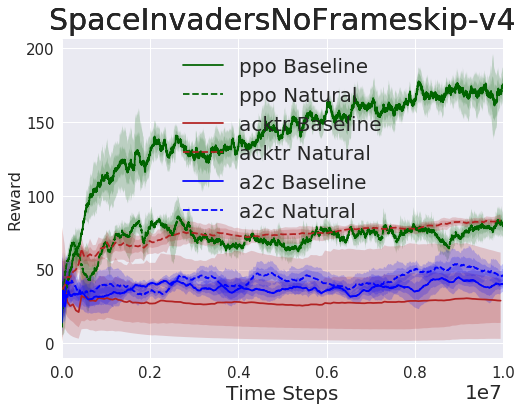}}
     \subfigure{\includegraphics[width=.24\textwidth]{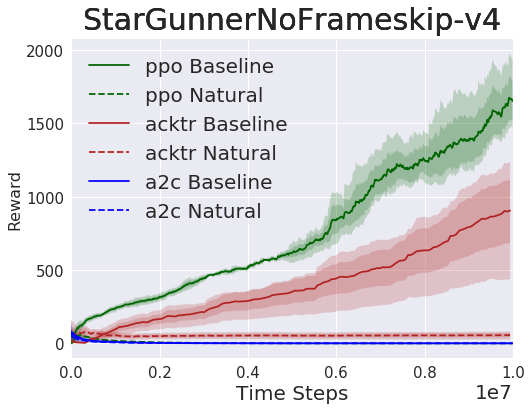}}
     \caption{Atari frames with natural video embedded as background, using \cite{pytorchrl} code implementation. Variance computed across 5 seeds.}
     \label{fig:pytorch_nat_atari_results}
 \end{figure*}

\end{document}